\documentclass[%
 reprint,
superscriptaddress,
 amsmath,amssymb,
 aps,
]{revtex4-1}

\usepackage{graphicx}
\usepackage{dcolumn}
\usepackage{bm}
\usepackage{hyperref}
\usepackage{cleveref}
\usepackage{natbib}
\usepackage{todonotes}

\begin{document}

\preprint{APS/123-QED}

\title{Road Network Deterioration Monitoring Using Aerial Images and Computer~Vision}

\author{Nicolas Parra-A}
    \email{nparraa@unal.edu.co}
\author{Vladimir Vargas-Calderón}
\affiliation{Laboratorio de Inteligencia Artificial y Computación de Alto Desempeño,
Human Brain Technologies, Bogotá, Colombia}
\affiliation{%
 Grupo de Superconductividad y Nanotecnología, Departamento de Física,\\
 Universidad Nacional de Colombia, Bogotá, Colombia}
\author{Herbert Vinck-Posada}%
\affiliation{%
 Grupo de Superconductividad y Nanotecnología, Departamento de Física,\\
 Universidad Nacional de Colombia, Bogotá, Colombia}
 
\author{Nicanor Vinck}
\affiliation{Departamento de Ingeniería Civil, Escuela Colombiana de Ingeniería Julio Garavito, Bogotá, Colombia}


\date{\today}

\begin{abstract}
Road maintenance is an essential process for guaranteeing the quality of transportation in any city.
A crucial step towards effective road maintenance is the ability to update the inventory of the road network.
We present a proof of concept of a protocol for maintaining said inventory based on the use of unmanned aerial vehicles to quickly collect images which are processed by a computer vision program that automatically identifies potholes and their severity.
Our protocol aims to provide information to local governments to prioritise the road network maintenance budget, and to be able to detect early stages of road deterioration so as to minimise maintenance expenditure.
\end{abstract}

\maketitle


\section{\label{sec:Intro}Introduction}

Highway and road infrastructure in densely populated cities is essential for their society and economy well-being. The quality of the roads is the basis to ensure fast and secure transportation of goods and people through the city. Road and highway quality, in the most primitive sense, is associated to the absence of potholes and cracks~\footnote{There are other aspects of road quality, such as the materials' durability, or higher-level road features such as proper sign posts, traffic lights programming, among others.}. If these are absent, the highway network can thrive towards fast transportation, meaning that vehicles can reach high speeds. Two outstanding advantages are derived from fast transportation: economy sectors that depend upon transportation will have less operational costs, positively impacting the economy; and, people will enjoy shorter-time commutes, increasing their quality of life. Furthermore, absence of potholes and cracks imply that vulnerable vehicles such as motorcycles are less prone to be involved in accidents due to poor road surface conditions. All these well-known advantages of having high-quality highways and roads make road maintenance a central task for city administration. As such, any step towards making road maintenance efficient, positively impacts the city budget, as well as the city administration's capacity to look after its highway network.

Since road maintenance is a central problem for traffic mobility in cities, several proposals have been made for pothole detection, including vibration measurements from vehicles, 3D reconstruction of the road through laser imaging and image processing methods~\citep{kim2014review,dhiman2020,SHTAYAT2020629}. The latter is the most prominent research branch for pothole detection systems, providing a two-fold advantage: first, collecting quality images is now easy due to the availability and low-cost of high-resolution cameras; and second, image processing methods have benefited in the last decade from deep learning for computer vision research.

Image collection has mainly focused on cameras mounted on passenger vehicles or specialised vehicles. With such images, different image processing methods for pothole detection have been used, such as pixel/segments clustering, histogram shape-based thresholding, disparity maps for stereo-vision images, among others~\citep{buza2013pothole,vigneshwar2016,nienaber2015detecting,lee2019,lo2015,KOCH2011507,lee2018,jo2015,shuai2016,fan2020,akagic2017,pesavento2020,OUMA2017196,liyaqi2018}. With the advent of deep learning, neural network-based algorithms have flourished in the pothole detection literature as well, particularly convolutional neural networks~\citep{chen2020,wanli2021,shaghouri2021realtime,ukhwah2019,dhiman2020,fan2021graph}.

Yet another improvement in image collection is achieved when using unmanned aerial vehicles (UAVs) such as drones for capturing images of the roads. UAVs allow covering larger areas in less time compared to road vehicles, yielding faster and cheaper quality image collection~\citep{ellenberg2014}. For this reason, they have started to be used in some studies for pothole detection schemes that use both classical image processing methods~\citep{pehere2020detection,SAAD2019647,leonardi2019,Fan2019CVPRWorkshops,tan2019,ellenberg2014,lee2019uav,leonardi20193d,zhang2008uav,dobson2013,Zhang2019study}, as well as deep learning methods~\citep{silva2020yolo,grzywinski2018,ZHU2022103991,pan2018,furusho2019,hassan2021,kotian2019unmanned}.

In this work, we present an inventory protocol that allows a low-cost and efficient way to assess the deterioration state of roads. We exemplify this protocol in a particular region of Bogotá, Colombia. The protocol consists of capturing aerial images of the highway network using drones, which are fed into a computer vision program that identifies potholes and classifies their severity. During the realisation of the project reported in this paper, other similar inventory protocols have been proposed and deployed in research stages across the world~\citep{ZHU2022103991,silva2020yolo}. Our method can serve as a near-real-time inventory that can be used by the city administration to make decisions about prioritising roads to be intervened for maintenance. 

Even though the inventory protocol that we present is useful for any city, it is particularly applicable for large cities such as Bogotá with a massive demand for roads due to the large number of users. Almost 7.5 million people inhabit Bogotá~\citep{dane2021}. A total of 2.4 million vehicles use the city roads~\citep{movilidadcifras2017}, which extend for 14.200 kilometres~\citep{mallavial2019}. The traffic in Bogotá is ranked as the 8th worst in the world in 2021 (it was the worst in the world in 2020)~\citep{pishue_2021}, with people losing an average of 94 hours in traffic per year. We emphasise that these hours are lost due to high traffic, but also due to poor road conditions.

This paper is organised as follows. \Cref{sec:Collection} explains the data set of collected street images from one of the regions with more road damage in Bogotá. \Cref{sec:Methodology} presents the methodology used to build a computer vision program for pothole identification and classification of their severity.
\Cref{sec:Results} shows the main results, such as quality metrics of the computer vision program as well as distributions of pothole density and severity.
Finally, \cref{sec:Conclusions} concludes the paper.

\section{\label{sec:Collection}Image collection}

\begin{figure}
    \centering
    \includegraphics[width=\columnwidth]{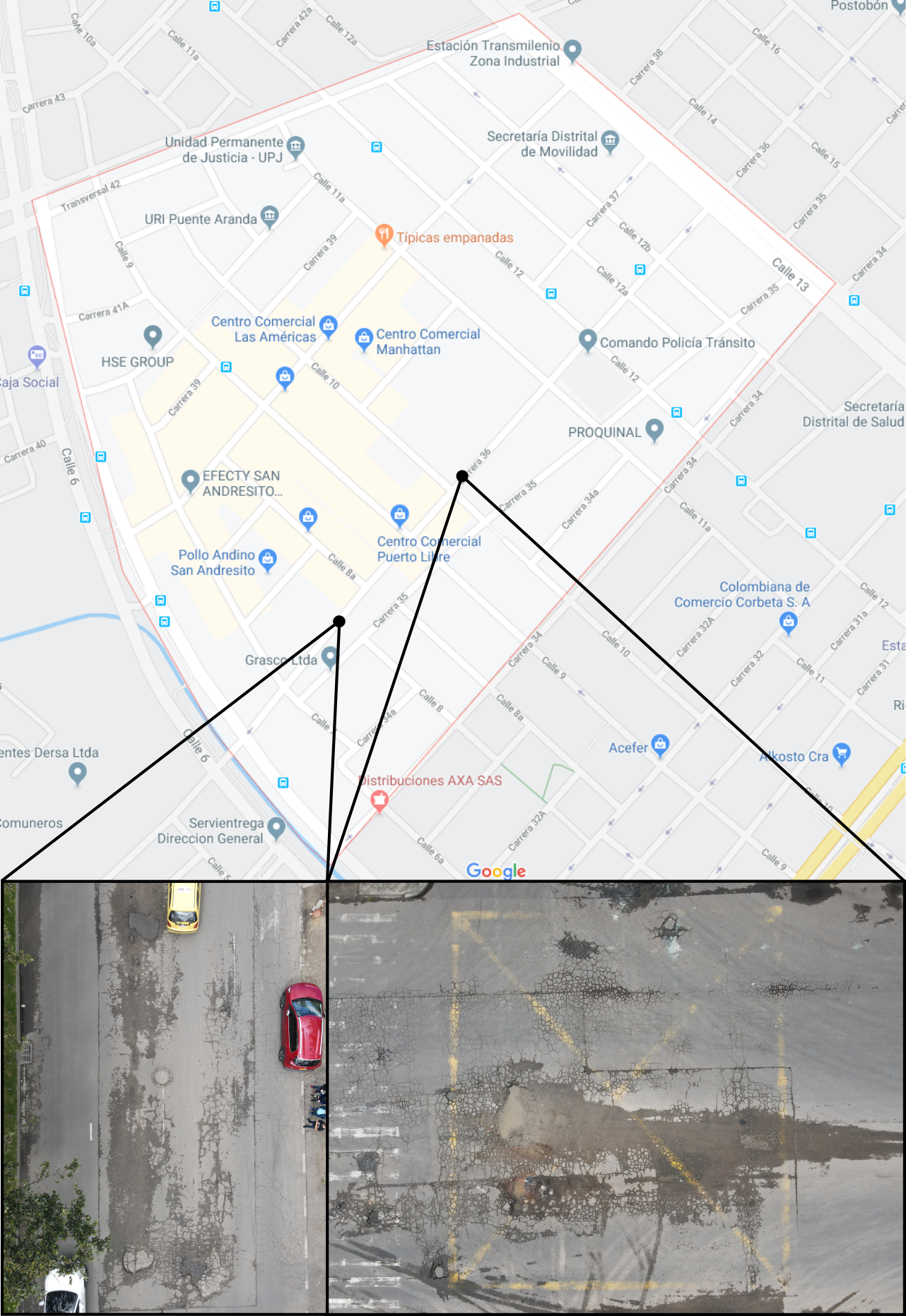}
    \caption{Map of the Industrial Zone in Bogotá, located at latitude 4.61600 and longitude -74.10096.}
    \label{fig:zona_industrial}
\end{figure}

We collected aerial images from an area in Bogotá that famously suffers from heavy goods traffic and low road maintenance, called the Industrial Zone (see~\cref{fig:zona_industrial}). The poor condition of many of the Industrial Zone's roads makes this area a particularly good zone to test pothole detection algorithms because of the diversity of shapes, sizes and severity of potholes.

Images were captured by a DJI Mavic 2 drone, which was flown in manual mode over the specified region because there was danger of colliding with some high structures and electricity cables.
Nonetheless, commercial software such as DroneDeploy~\footnote{\url{https://www.dronedeploy.com/}} offers programmed flights plus the ability to reconstruct 3D surfaces using stereo-graphic techniques, which are becoming popular in both research an industry.
\citet{romero2020} showed that quality images to resolve all types of road distress can be achieved with heights between 10 and 15 metres above ground.

\begin{figure}
    \centering
    \includegraphics[width=\columnwidth]{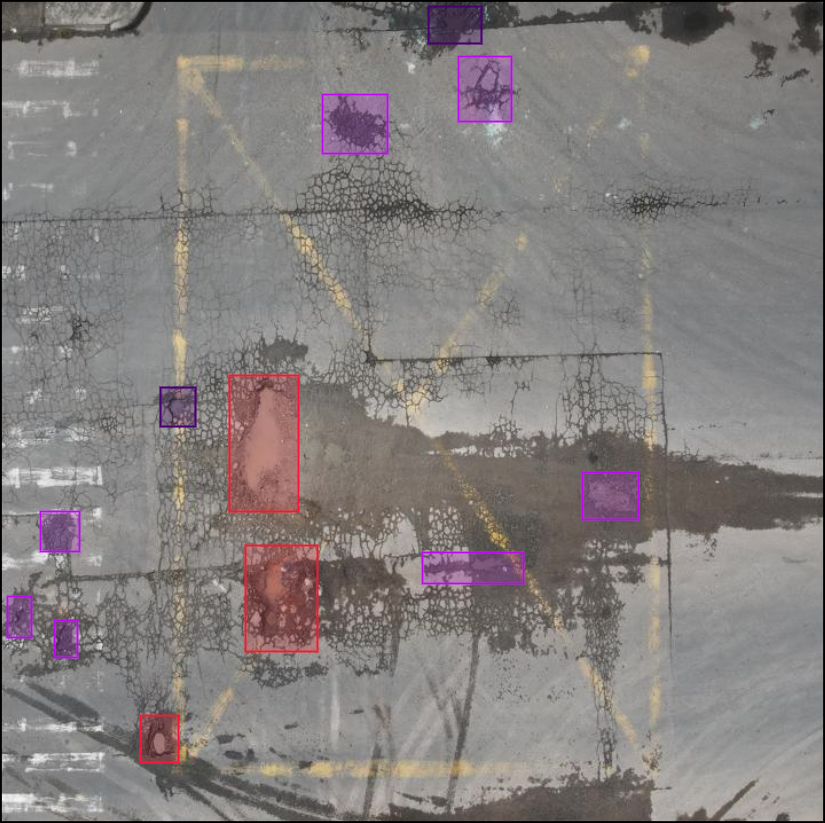}
    \caption{Example of potholes annotation with the \url{https://www.makesense.ai/} tool. Rectangles in light purple correspond to low severity potholes, dark purple to middle and red to high severity potholes.}
    \label{fig:anotacion}
\end{figure}

A total of 149 images were taken in streets where different types of potholes and cracks were present.
Additionally, 62 images were rotated in arbitrary angles to increase the data set size so that the computer vision program could learn more general potholes shapes.
After of this process our data set has 211 images.
From these images, we manually identified and annotated 1.777 potholes of different severity (low, middle and high).
\Cref{fig:anotacion} shows an example of an annotated image.

\section{\label{sec:Methodology}Methodology}

The task of identifying and classifying the severity of the images is solved through the you only look once (YOLOv3) algorithm~\citep{Redmon_2016_CVPR}.
Before delving in how the algorithm works, it is important to highlight that it is made for rapid inference, usually keeping prediction feasible in real-time.
Moreover, it can be light enough to run in mobile hardware such as a Raspberry Pi~\citep{asmara2020prediction}, which further enables the possibility of feeding structured information from already processed images to a central system instead of feeding heavy images to perform inference in a centralised computer.

YOLO is essentially composed of a neural network model that takes as input an image of a given size, and outputs a prediction grid.
This is done by splitting the input image in a given grid of $S\times S$ patches.
Each patch or cell will have the task of predicting $B$ bounding boxes along with associate confidence scores, which tell the model's confidence that the corresponding bounding box contains an object.
Yet another dimension can be added by not simply predicting where an object is, but what kind of object it is.
The input image goes only once through the neural network, which is why the model was given the name ``you only look once''.

Each bounding box is a prediction vector composed of the components $x, y, h, w, c$, where $(x,y)$ are the pixel centre coordinates of a box of height $h$ and width $w$, and
\begin{align}
    c = \text{Pr}(\text{Object})\times\text{IOU},
\end{align}
indicates the confidence that there is an object in the bounding box, and IOU is the intersection over union, which quantifies the confidence that the bounding box and the actual object coincide.
If the confidence values do not surpass a certain threshold, it is assumed that the bounding box does not contain an object.
Furthermore, for each patch, a vector of probabilities $\text{Pr}(\text{Object}=\ell)$ is computed to assign a class $\ell$ out of $C$ classes to the object identified in the patch, if such an object is identified.
Overall, there are a total of $S^2(5B+C)$ output values for the prediction of the YOLO neural network.

The neural network is trained by minimising a sum of squares loss, typical of regression problems, where predicted and true values are subtracted, squared, and then summed.
Since each patch has $B$ bounding boxes, it is assumed that a patch can only hold one object.
Thus, during training, the predicted parameters that characterise the box ($x,y,h,w,c$) of a given object are those of the bounding box with higher IOU value.

\section{\label{sec:Results} Results}

\begin{figure}
    \centering
    \includegraphics[width=\columnwidth]{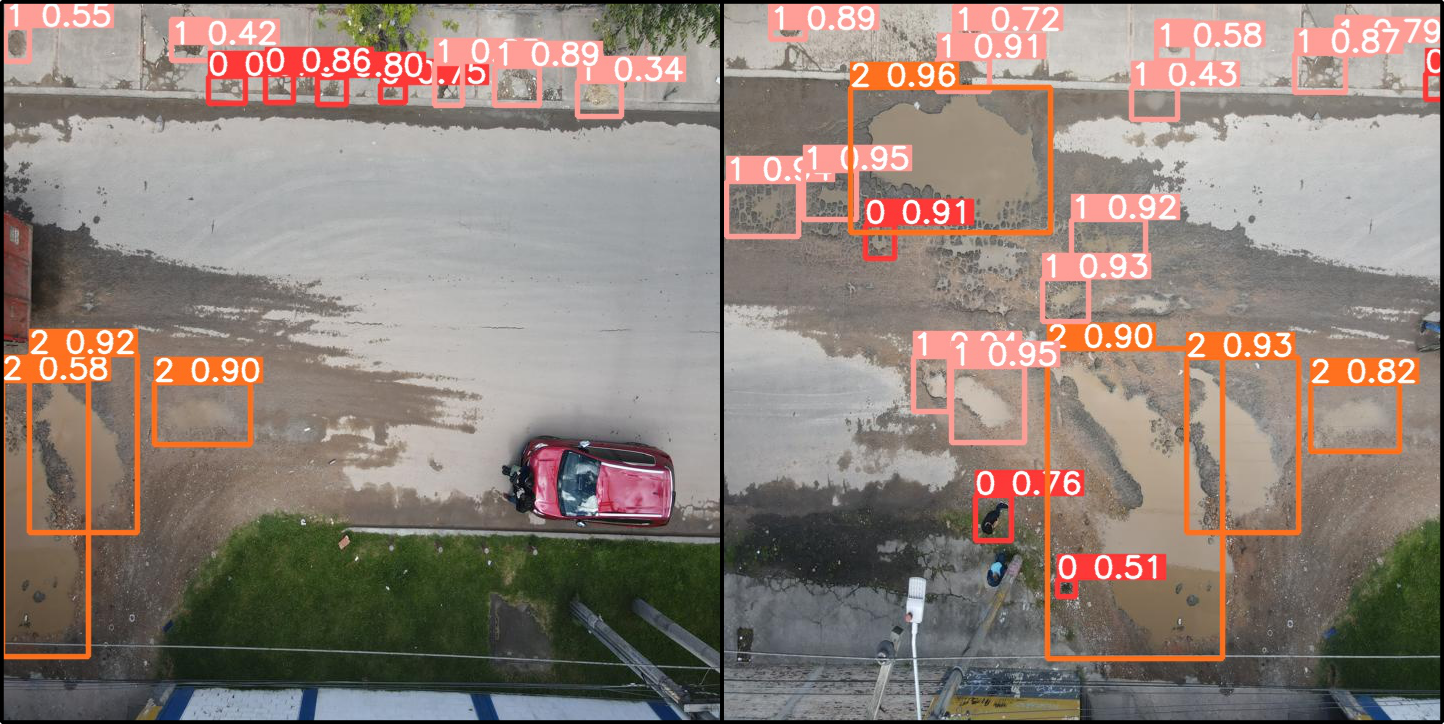}
    \caption{Example of predictions for the trained YOLO model.
    In each bounding box we can see a integer number and a float number.
    The integer number corresponds to the pothole's severity,
    zero for low, one for middle and two for high severity.
    The float indicates the probability that the bounding box contains a pothole, i.e., this is the confidence.}
    \label{fig:prediccion}
\end{figure}

The dataset was split so that 80\% of the images were used to train YOLO, and 20\% of them were used for testing.
The achieved IOU metric in the test set was 0,59.
Fig \ref{fig:prediccion} shows an image annotated by YOLO, which shows how YOLO is able to detect potholes of different severity.

It is important to remark that transversal and longitudinal cracks are also an important source of road deterioration, which have been not included in the current study.
This means that we have not annotated these sources.
Nonetheless, YOLO is able to generalise the visual features of potholes to those of other types of road deterioration as shown in Fig \ref{fig:prediccion}.

\begin{figure}
    \centering
    \includegraphics[width=\columnwidth]{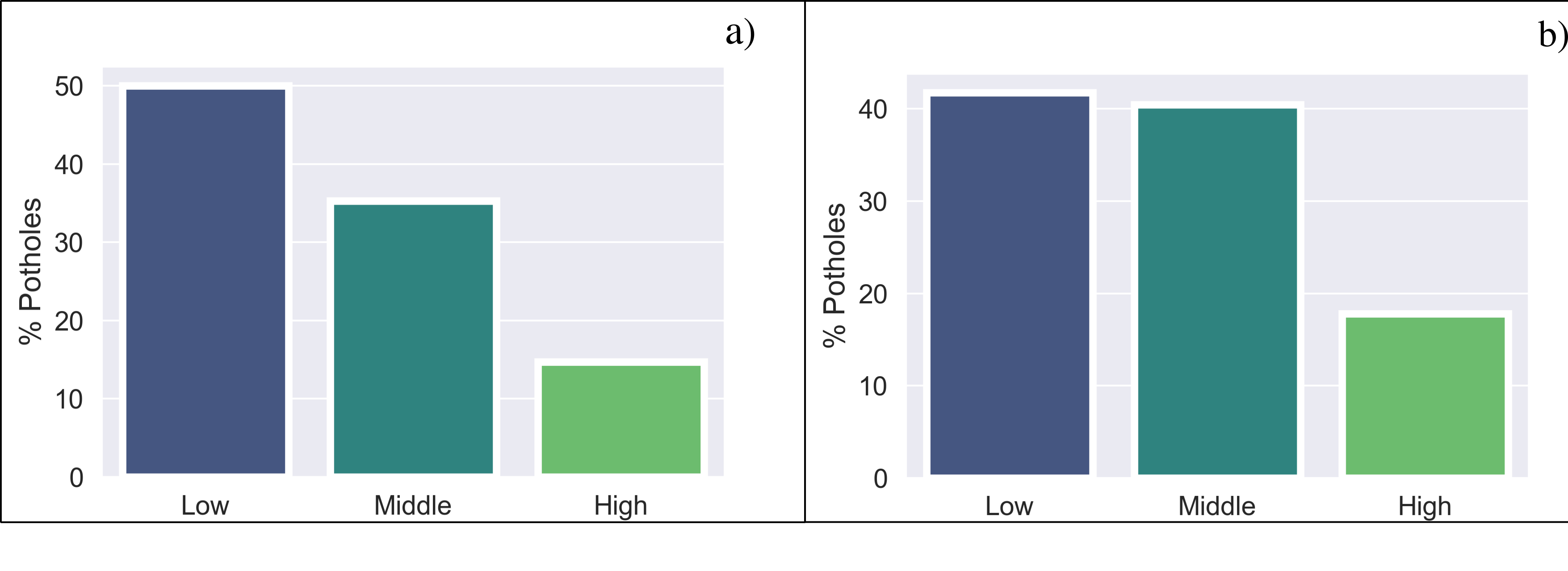}
    \caption{a) pothole distribution in train set, b) pothole distribution in predicted data for the low, middle and high severities.}
    \label{fig:dist_potholes}
\end{figure}

Density of pothole severities in the train set and the test set after inference are shown in~\cref{fig:dist_potholes}.
It is clear that the severity distribution in the test set after inference is not the same as in the annotations found in the train set.
This could be improved with further annotation and with an increase of image variability by taking pictures in other conditions.
Furthermore, some potholes that need to get discarded are those that do not correspond to the road itself but to the sidewalks: this is left for future work.
However, since the aim of our project is to provide useful information for road maintenance, it can be argued that sidewalks are also part of the road infrastructure.
Therefore, having information about these potholes can be important.
We propose to annotate them with a different label, or set of labels, if their severity also is to be assessed.

\section{\label{sec:Conclusions} Conclusions}

We have proposed a protocol for maintaining an inventory of the quality of road networks.
This protocol consists of using unmanned aerial vehicles to collect images of the roads.
These images are then processed by a computer vision program based on the YOLO algorithm for object detection.
In particular, potholes and their severity are automatically detected by this algorithm.

Our protocol provides a scalable and low-budget strategy for maintaining road network inventories, which can help in prioritising the road maintenance budget.
Further, if the roads are overall in good condition, being able to identify early deterioration signs in the roads can further decrease maintenance expenditure, as simpler and cheaper road maintenance techniques can be used to correct the road damage.

We highlight a couple of drawbacks of collecting images with UAVs.
Our study was conducted with manual control of the UAVs, but a faster and more scalable way is through automated flights.
However, automated flights imply inherent risks such as having erroneous instructions that can lead to UAVs crashing on high structures in the cities that were not foreseen when designing the flight plans for automatic flights.
Also, UAVs not being taken care of by a human pilot on land can make the UAVs prone to being stolen.
Furthermore, there are serious limiting conditions that hamper the process of image collection, namely traffic and weather.
Since images are aerial, if too much traffic is present, it is obvious that the road cannot be photographed.
Moreover, rainy conditions will prevent UAVs from flying.
Also, if images are collected after raining, the water itself can be problematic since it can hide several visual features of the potholes.
\section*{Acknowledgements}
We thank Juan S. Rojas for his help in annotating parts of the aerial image dataset. N. P.-A, V. V.-C. and H. V.-P. thank project HERMES-48528 for financial support.

\bibliography{apssamp}

\end{document}